\documentclass[letterpaper, 10 pt, conference]{ieeeconf}  

\IEEEoverridecommandlockouts                              

\usepackage{graphicx, subfigure}
\usepackage{multirow}
\usepackage{xcolor}

\usepackage{graphicx}
\usepackage{xcolor}
\usepackage{fancyhdr}

\title{\LARGE \bf
Multi-Instance Multi-Label Learning for Gene Mutation Prediction in Hepatocellular Carcinoma
}

\usepackage{etoolbox}
\makeatletter
\patchcmd{\@makecaption}
  {\scshape}
  {}
  {}
  {}
\makeatother

\author{Kaixin Xu$^{*2}$, Ziyuan Zhao$^{*1}$, Jiapan Gu$^{1}$, Zeng Zeng$^{\dagger 1}$\\
Chan Wan Ying$^{3}$, Lim Kheng Choon$^{4}$, Thng Choon Hua$^{3}$, and Pierce KH Chow$^{5}$
\thanks{*The first two authors contribute equally in this work. $^{\dagger}$Corresponding author. The work was supported by Pre-GAP Grant, Singapore (Grant No. ACCL/19-GAP023-R20H) and Singapore-China NRF-NSFC Grant (Grant No. NRF2016NRF-NSFC001-111). $^{1}$Institute for Infocomm Research (I2R), Agency for Science, Technology and Research (A*STAR), Singapore. $^{2}$National University of Singapore, Singapore. $^{3}$National Cancer Centre, Division of Oncologic Imaging, Singapore. $^{4}$Singapore General Hospital, Department of Vascular and Interventional Radiology, Singapore. $^{5}$National Cancer Centre, Division of Surgery and Surgical Oncology, Singapore. }
}

\begin{document}

\maketitle
\thispagestyle{empty}
\pagestyle{empty}

\thispagestyle{fancy}
\fancyhead{}
\lfoot{}
\lfoot{\scriptsize{Copyright 2020 IEEE. Published in the 2020 42nd Annual International Conference of the IEEE Engineering in Medicine and Biology Society (EMBC), scheduled for July 20-24, 2020 at the Montréal, Canada. Personal use of this material is permitted. However, permission to reprint/republish this material for advertising or promotional purposes or for creating new collective works for resale or redistribution to servers or lists, or to reuse any copyrighted component of this work in other works, must be obtained from the IEEE. Contact: Manager, Copyrights and Permissions / IEEE Service Center / 445 Hoes Lane / P.O. Box 1331 / Piscataway, NJ 08855-1331, USA. Telephone: + Intl. 908-562-3966.}}
\rfoot{}

\begin{abstract}

Gene mutation prediction in hepatocellular carcinoma (HCC) is of great diagnostic and prognostic value for personalized treatments and precision medicine. In this paper, we tackle this problem with multi-instance multi-label learning to address the difficulties on label correlations, label representations, etc. Furthermore, an effective oversampling strategy is applied for data imbalance. Experimental results have shown the superiority of the proposed approach.
\newline

\indent \textit{Clinical relevance}— The proposed framework applies multi-instance multi-label learning for gene mutation detection in HCC, which can be implemented to provide clinical diagnostic and predictive oncology services.
\end{abstract}

\section{INTRODUCTION}


Hepatocellular carcinoma (HCC) is the most common complication and the main cause of death among patients with cirrhosis~\cite{davis2008hepatocellular}. As the primary tumor of the liver, the diagnosis of HCC relies on abdominal imaging, such as computed tomography (CT), in which, various image traits (biomarkers) are identified for further analysis. In common, a liver biopsy is performed to support the microscopic and molecular analysis of HCC, which provides much prognostic and therapeutic information for designing targeted therapy and precision medicine. Once the global gene expression in HCC is obtained, targeted drugs can be applied to control the specific gene expression in HCC. But the invasive biopsy is destructive and risky. In recent years, a potential non-invasive approach ``radiogenomics"~\cite{rutman2009radiogenomics}  has been rapidly developed to bridge the gap between radiomics and genomics. Many research work have demonstrated that imaging features are correlated with cancer genomics~\cite{segal2007decoding, aerts2014decoding}.

\begin{figure*}
    \centering
    \includegraphics[width=0.85\textwidth]{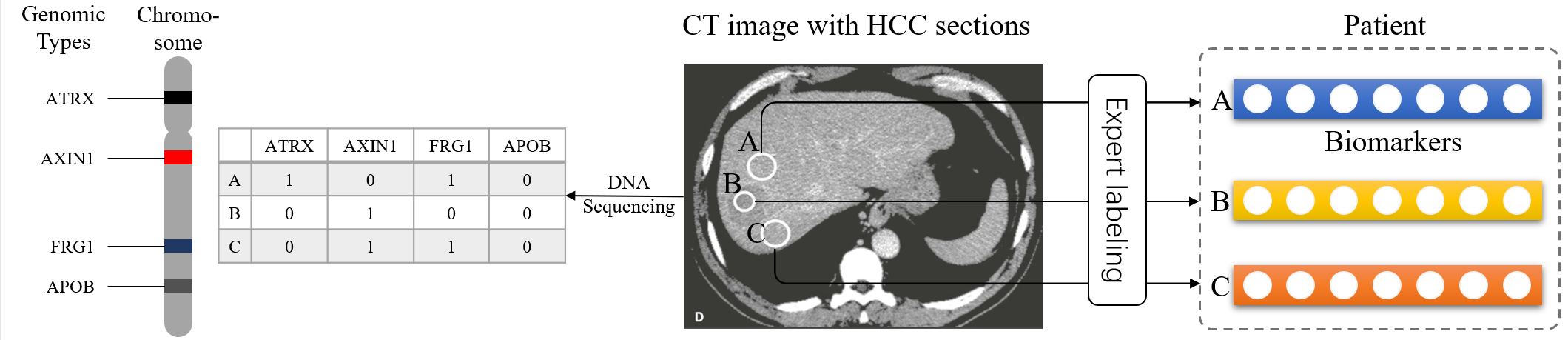}
    \caption{The transformation from HCC gene types diagnosis program to machine classification. As a demonstration, the relative locations of different gene types on the left-most chromosome may not be in the exact order on DNA string, or can even be on different chromosomes~\cite{c11asd}.}\label{fig:overall}
\end{figure*}

Exploring the associations, between image traits (biomarkers) and gene types of tumors, can be facilitated using supervised learning methods~\cite{zhao2019bira}. More specifically, a classifier can be trained using a dataset comprised by biomarkers and labels of gene types, where each tumor sample is probable to relate to more than one gene mutations. As demonstrated in Fig.~\ref{fig:overall}, biomarkers are collected through robust image feature identification performed by expert radiologists. Utilizing the annotated biomarkers, a set of classifiers can be trained to determine gene mutations. In this scenario, the problem is transformed into a multi-label learning (MLL) problem.

However, the development of MLL in gene mutation detection is critically impeded by the following challenges. First, the intra-tumour heterogeneity in HCC~\cite{zhai2017spatial} results in the difference of gene types in different locations of tumors, so that some biomarkers do not correspond to the assigned mutation types. This scenario makes MLL approaches difficult to train for mutation detection in HCC. Second, biomarker determination is laborious, costly, error-prone and always suffers from high intra / inter-observer variability, therefore,  some biomarkers may be missed or mislabeled.


In this paper, we propose a Multi-Instance Multi-Label learning (MIML) framework, a variant of the conventional MIML approach~\cite{c3}, to alleviate the above problems, where we use multiple instances for robust multi-label classification. We evaluate the proposed method on a large-scale public dataset and a genomics dataset collected from multiple hospitals based on Singapore. The experimental results demonstrate that our framework outperforms the conventional approaches. Besides, we applied oversampling strategy in the genomics data on top of MIML, further improving the performance of gene mutation detection in HCC.

\section{RELATED WORK}
\label{sec:mil}

\begin{figure}[h]
    \centering
    \subfigure[MISL]{
    \includegraphics[width=0.2\textwidth]{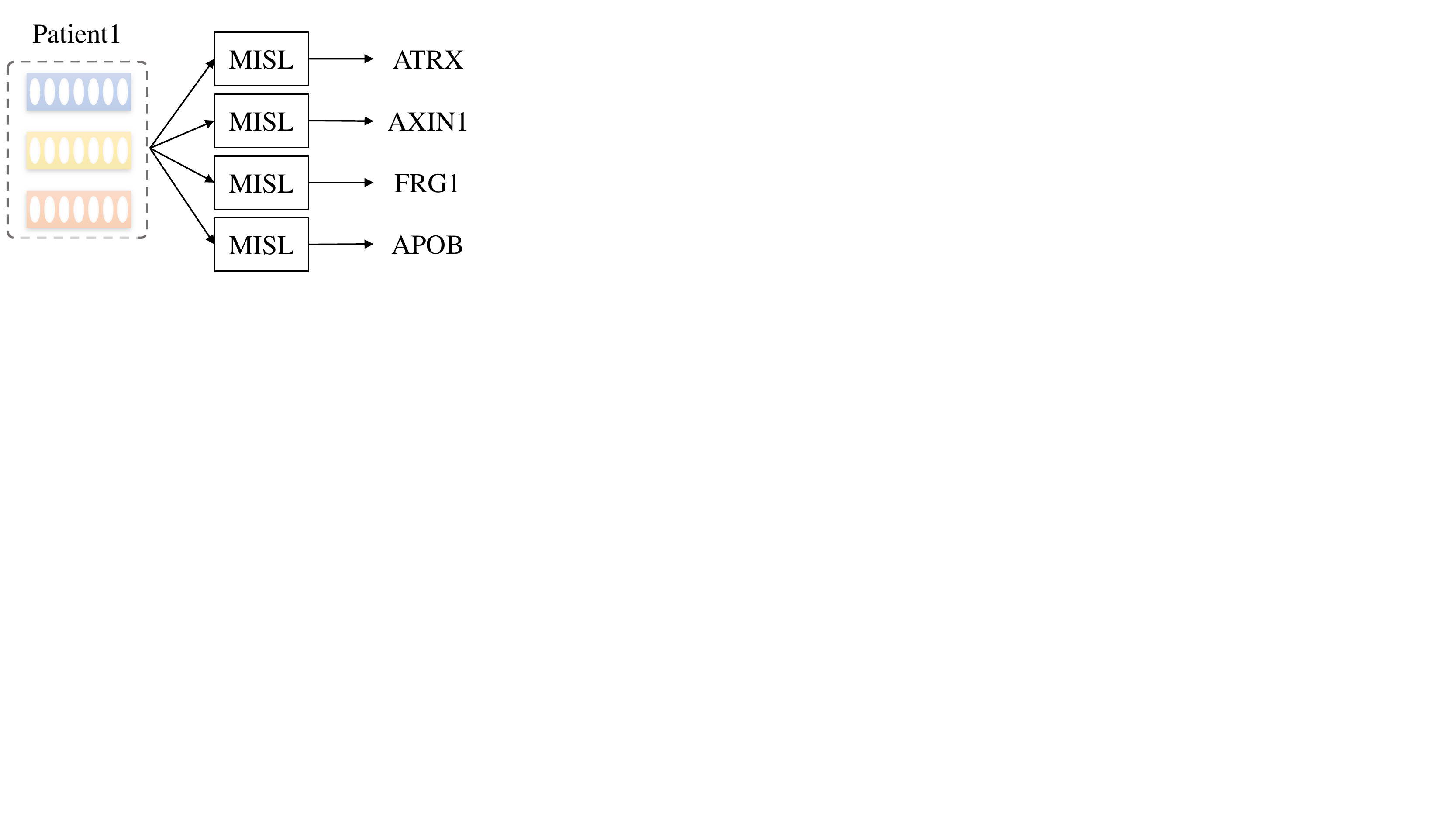}\label{fig:misl}
    
    }
    \subfigure[SIML]{
    \includegraphics[width=0.2\textwidth]{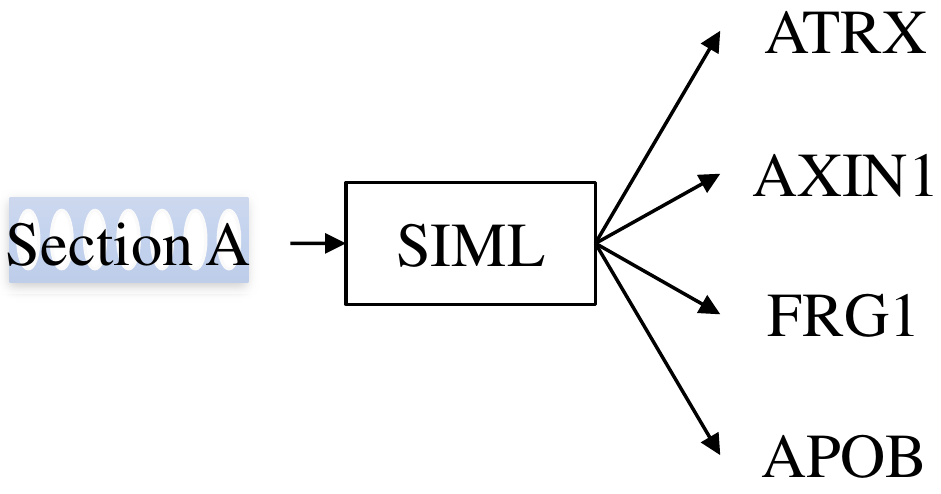}\label{fig:siml}
    }
    \subfigure[MIML]{
    \includegraphics[width=0.2\textwidth]{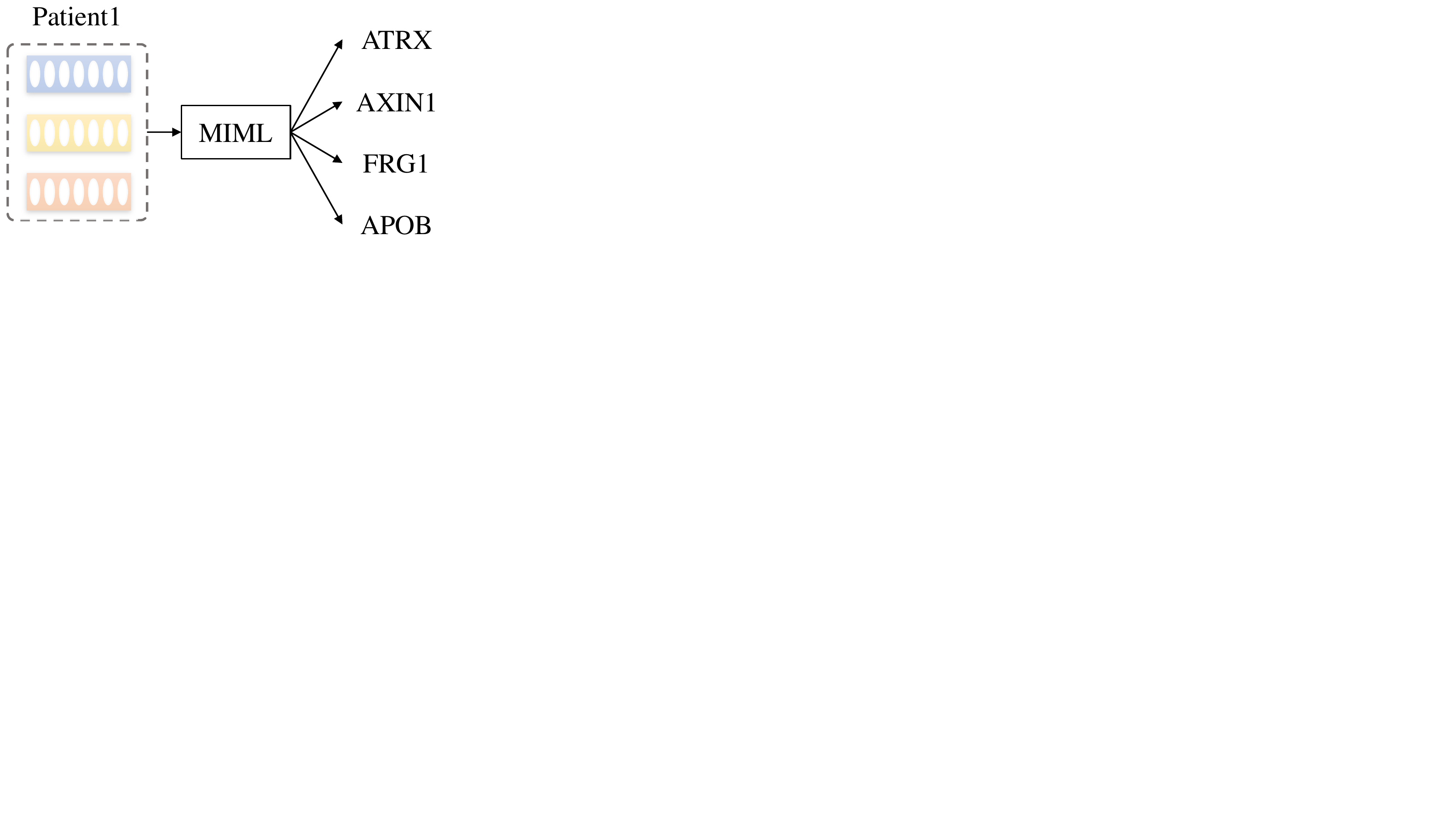}\label{fig:miml}
    }
\caption{Explanation of different learning strategies: (a) Multi-Instance Single-Label Learning (MISL) (b) Single-Instance Multi-Label Learning (SIML) (c) Multi-Instance Multi-Label Learning (MIML)}\label{fig:sl}
\end{figure}

In multi-instance learning (MIL), the classifier learns the mapping $f_{MIL}: 2^{\mathcal{X}} \rightarrow \{\pm1\}$, where the training dataset is a set of labelled bags (e.g. mutation present/absent in patient) $X=\{X_i\}_{i=1}^n$, each containing multiple instances (biomarkers in different locations) $X_i = \{\mathbf{x}_j^{(i)}\}_{j=1}^{n_i}$ (see Fig.~\ref{fig:misl}). Compared to single instance learning (SIL), MIL learns from the bag as a whole instead of individual instances, which leads to looser requirements of the performance at bag level than instance level . In other words, it focuses on the gene mutations happened at coarse-grained level (the tumor) rather than fine-grained level (locations of gene mutations in the tumor). Bunescu~\textit{et al.}~\cite{c4} tackled the MIL problem using semi-supervised learning~\cite{zhao2019semi}, in which positive (negative) bags were regarded as unlabeled (labeled) samples, then constrained instance SVM like sMIL was performed.


In multi-label learning (MLL), the classifier learns the mapping  $f_{MLL}: \mathcal{X} \rightarrow 2^{\mathcal{Y}}$, where a set of labels $\mathcal{Y}$ are assigned to the instance $X$, and each label $y \in \mathcal{Y}$ has chance to appear in the output of the prediction $\mathcal{\widehat{Y}}$ (see Fig.~\ref{fig:siml}). Classifier chains~\cite{c5, KDD2018} is one of the standard MLL methods, in which, conditional dependence between a label and its predecessors is modeled. Other approaches like adopting label Power Set (PS) transformation, also take label correlations into account~\cite{multitarget}.



In multi-instance multi-label learning (MIML), the classifier learns the mapping from a bag of instances to a set of labels, as shown in Fig.~\ref{fig:miml}, which is a combination of MIL and MLL. Zhou~\textit{et al.}~\cite{c3} formulated a joint MIML framework for multi-instance multi-label samples and proposed two approaches termed as MIMLBoost and MIMLSVM, both of which make discrimination at bag level (under Bag-Space paradigm~\cite{c6}). Gene mutation determination benefits from both merits of MIL and MLL under MIML framework. On one hand, targeted drugs should be applied to a patient with the gene mutations even if the gene is only mutated on a small part of the tumor, in which, multi-instance inference is suitable. On the other hand, MLL is capable of modeling potential relationships among gene mutations. 



\begin{table*}[th]
\caption{Experimental results on the scene classification dataset with different regularization coefficients (C). $\uparrow$ indicates that the metrics are positively correlated with the model performance, on the contrary, $\downarrow$ means that the smaller metrics are better. Values before $\pm$ and after are the averages and the standard variances in cross validation respectively.}\label{fix_ccrit}
    \centering
    \begin{tabular}{c|c||c|c|c|c|c}
        \hline
        \multicolumn{7}{c}{MLL criterion: T-Criterion, k-medoids's $k=300$} \\ \hline
        Method & C & Hamming Loss$^{\downarrow}$ & EM$^{\uparrow}$ & Rank Loss$^{\downarrow}$ & Coverage$^{\downarrow}$ & AP$^{\uparrow}$ \\ \hline
        \multirow{3}*{MIMLSVM} & $10^{-4}$ & $0.38\pm 0.005$ & $0.13\pm 0.18$ & $0.43 \pm 0.11$ & $2.37\pm 0.37$ & $0.33\pm 0.12$ \\ \cline{2-7}
        ~ & $10^{-3}$ & $0.34\pm 0.08$ & $0.19\pm 0.14$ & $0.42\pm 0.03$ & $2.38\pm 0.37$ & $0.32\pm 0.12$ \\ \cline{2-7}
        ~ & $10^{-2}$ & $0.34\pm 0.09$ & $0.19\pm 0.17$ & $0.42\pm 0.14$ & $2.36\pm 0.43$ & $0.33\pm 0.16$ \\ \hline
        \hline 
        \multirow{3}*{Ours} & $10^{-4}$ & $\mathbf{0.21\pm 0.005}$ & $0.42\pm 0.03$ & $0.24\pm 0.01$ & $1.79\pm 0.04$ & $0.55\pm 0.01$ \\ \cline{2-7}
        ~ & $10^{-3}$ & $0.21\pm 0.007$ & $\mathbf{0.43\pm 0.03}$ & $\mathbf{0.22\pm 0.02}$ & $\mathbf{1.75\pm 0.06}$ & $\mathbf{0.56\pm 0.02}$ \\ \cline{2-7}
        ~ & $10^{-2}$ & $0.21\pm 0.01$ & $0.40\pm 0.02$ & $0.23\pm 0.01$ & $1.77\pm 0.04$ & $0.53\pm 0.02$ \\ \hline
    \end{tabular}

\end{table*}


\section{METHODOLOGY}

\subsection{The proposed MIML framework}

We propose a framework that is partly extended from MIMLSVM~\cite{c3}, in which, the multi-instance samples $X_{i} \in R^{n_{i} \times n_{f e a t}}$ are transformed into single vector representation $z_i$ using constructive clustering~\cite{c8}, then an MLL classifier based on label-independent single-instance SVMs is learned. In other words, an MIML function $f_{MIML}: 2^{\mathcal{X}} \rightarrow 2^{\mathcal{Y}}$ is converted into MLL by this transformation. An Embedded-Space strategy is then adopted in the MIL part, which compresses the structural information on instances within a random-size bag into a fixed-size single vector so that instance-level correlations are preserved. However, this strategy does not consider the label correlations, and gene mutations cannot be treated independently in our case. Therefore, in our framework, the SVMs in MIMLSVM and the GA-PartCC~\cite{c7} Classifier chains models are cascaded sequentially, with an optimization of the order of the chain using Genetic Algorithm. This allows the label set to form a partial chain, in which, the absent labels in the chain are regarded individually. The training process of the framework can be summarized as follows:

\begin{itemize}
    \item Given the original dataset $X=\{X_i\}_{i=1}^{n_{bag}}, (X_i = \{\mathbf{x}_1^{(i)}, \mathbf{x}_2^{(i)}, \dots, \mathbf{x}_{n_{i}}^{(i)}\})$. Here $n_i$ denotes the cardinality of $i$th bag, and $\mathbf{x}_j^{(i)}$ denotes the $j$th instance inside the $i$th bag. Each instance consists of $\{x_{1j}^{(i)}, x_{2j}^{(i)}, \dots, x_{kj}^{(i)}\}$ where k is the number of features.
    \item Constructive Clustering: apply k-mediods clustering of the bags using Hausdorff distance as the bag-level distance metric, transform the original dataset into a $Z={z_i}^{n_{bag}} \in {R}^{n_{bag} \times k}$, each bag $X_i$ is transformed into $z_i = \{dist(X_i - M_j) \mid 1 \le j \le k\}$.
    \item The first population in GA are chains randomly generated with different lengths and orders, but also includes an empty chain and a fully sequential chain.
    \item Train the population through sufficient generations, following the training paradigm described in \cite{c7} including crossover and mutation operations, selection mechanism (tournament) and the definition of fitness function. 
    \begin{itemize}
        \item During the sequential training along the chain $L_c=\{l_j\}^{n_c}$, the predecessor estimation of label relevance $\hat{y}_{j-1} \in [0, 1]$ is appended to the last column of $Z$, then feed the extended array of shape $(n_{batch}, n_{feat} + j - 1)$ to the next SVM classifier to predict $\hat{y}_j$. Labels that are not in the chain are predicted from $Z$ without previous predictions.
        \item Apply T/C-criterion~\cite{c8} on $\widehat{Y}^{(i)}$ to convert probabilistic estimation into binary prediction.
    \end{itemize}
    \item When the population stops iterating, pick the individual chain with the best fitness function as the ideal chain for later inference and evaluation.
\end{itemize} 

The cardinality of the bag in multi-instance dataset is nonuniform across different bags, thus the distance metric should be among subset of instances in the metric space instead of single instances. In practice, we use Hausdorff distance to measure the distance between two bags that contains multiple instances. 

\subsection{Balancing multi-instance dataset}
\label{subsec:oversample}


Compared to single-instance learning, where the imbalance of datasets occurs at instance level, the imbalance of multi-instance datasets occurs at both instance level and bag level. This could critically affect the performance of our MIML framework. Due to the nature of the liver tumor, the genomics dataset is extremely imbalanced with many positive labels. Assuming an HCC dataset balanced at instance level, the possibility of the HCC instances related to a specific mutation is $Prob_{ins}(y=1)=0.5$. Since patients will be marked as positive on the gene mutation if the gene mutation occurs on any part (instance) of the tumor, the possibility of the patient marked as positive becomes $Prob_{bag}(y=1) = 1 - (1 - Prob_{ins}(y=1))^{n_i} > 0.5$ if $n_i >= 2$. The imbalance at bag level will be even worse if $Prob_{ins}(y=1)$ itself is larger than $0.5$.


Since the MIL, as stated in Section~\ref{sec:mil}, ignores the instance-level details, we suspect that the multi-instance datasets predominated by positive examples will result in the decision boundaries of classifiers to drift towards negative bags and overfit the training set. Consequently, a negative instances oversampling strategy is proposed as follows:

\begin{itemize}
    \item Given the training example set $T=\{(X_i, Y_i)\}_{i=1}^m$, where $X_i$ denotes a bag and $Y_i$ the bag-level label, extract the negative instance set $T^-=\{X_i \mid Y_i = 0, 1 \le i \le m\}$ and unravel it into instance-level negative set $\tilde{T}^- = \{\mathbf{x}_1^{(1)},\mathbf{x}_2^{(1)}, \dots, \mathbf{x}_{n_1}^{(1)}, \mathbf{x}_1^{(2)},\dots, \mathbf{x}_{n_m}^{(m)}\}$.
    \item Randomly generate negative instances from the pool $\tilde{T}^-$ and then compose them into bags. The sizes of each generated negative bags are also a randomly generated integer larger than 1.
\end{itemize}


The above strategy is only appropriate for datasets where the instances in each bag can be regarded as independently drawn from instance space. This is applicable to genomics dataset because the combinations of biomarkers in different tumors of any patient are unrelated to each other, {\textit{i}.\textit{e}.}, $P(\mathbf{x}_{1:n_i}) = \prod_{i=1}^{n_i}{P(\mathbf{x}_i)}, \forall \mathbf{x}_{1:n_i} \in X_i$.

\section{EXPERIMENTS}
\subsection{Datasets and implementation}
\label{subsec:es}

The proposed framework is evaluated on two datasets. One is a public scene classification dataset collected from~\cite{c3}, which consists of 2000 bag samples in total, each bag of which contains exactly 9 instances with 15 attributes extracted from original RGB natural images and 2000 ground-truth labels, describing 5 scene categories including desert, sun, and sunset. Another dataset is collected from multiple hospitals located in Singapore, which contains biomarker sequences and genomics information from 27 patients with the approval of the Institutional Review Board, which has over 100 instances. Nine binary biomarkers in the genomics data, such as ``Arterial enhancement and Washout", were labeled as 0 / 1 by radiologists based on CT scans of patients.


Our method was implemented in Python based on scikit-learn. We conducted 5-fold cross validation experiments on the scene dataset, while 4-fold cross validation was taken on the genomics dataset. We compared our results with the baseline MIMLSVM model. Most of the parameter configurations of the baseline model followed~\cite{c3}, except in SVM, we used polynomial kernel instead of Gaussian kernel, where we set the kernel coefficient $c=0$ and the degree $d=3$ in default. In the genetic algorithm, the population size is set to 10, tournament size to 3.


\subsection{Results and discussion}

To validate the performance of the proposed method, Hamming Loss (HL), Exact Match (EM), Average Precision (AP), etc., are selected as the performance metrics~\cite{c3, c7}. 

On the scene classification dataset, we compared the performances of several scores using different SVM coefficient selections. As shown in Table~\ref{fix_ccrit}, our methods significantly improve the scores from the MIMLSVM framework among all settings. Average Precision (AP), which shows the average accuracy over all labels, has a 30\% improvement from MIMLSVM. Even the Exact Match (EM), the most rigorous criterion, was also vastly elevated from around 20\% to over 40\%. Moreover, the statistics show that the results of our method can consistently perform better than the baseline, and most of the top scores are obtained when $C=10^{-3}$.

\begin{table}[htb]
\caption{Experimental results on the genomics dataset with regularization coefficient $C=0.1$ and different C-criterion (C-crit).  When C-crit$=0$, T-criterion will be applied.}\label{fix_c}
    \centering
    \begin{tabular}{c|c||c|c|c}
        \hline
        Method & C-crit & CL & HL$^{\downarrow}$ & AP$^{\uparrow}$ \\ \hline
        SISL & $0.0$ & / & $0.26\pm 0.02$ & $0.59\pm 0.07$ \\ \hline
        \multirow{5}{*}{MIMLSVM} & $0.0$ & / & $0.27\pm 0.02$ & $0.67\pm 0.05$ \\ \cline{2-5}
        ~ & $0.1$ & / & $0.26\pm 0.02$ & $0.73\pm 0.06$ \\ \cline{2-5}
        ~ & $0.2$ & / & $0.27\pm 0.02$ & $0.69\pm 0.04$ \\ \cline{2-5}
        ~ & $0.3$ & / & $0.25\pm 0.03$ & $0.71\pm 0.03$ \\ \cline{2-5}
        ~ & $0.4$ & / & $0.27\pm 0.07$ & $0.71\pm 0.03$ \\ \hline
        \multirow{5}{*}{Ours} & $0.0$ & $5.75\pm 7.5$ & $0.23\pm 0.02$ & $0.70\pm 0.02$ \\ \cline{2-5}
        ~ & $0.1$ & $3.25\pm 2.5$ & $\mathbf{0.23\pm 0.02}$ & $0.73\pm 0.036$ \\ \cline{2-5}
        ~ & $0.2$ & $3.75\pm 3.5$ & $0.24\pm 0.02$ & $0.72\pm 0.03$ \\ \cline{2-5}
        ~ & $0.3$ & $5.75\pm 7.5$ & $\mathbf{0.23\pm 0.02}$ & $\mathbf{0.74\pm 0.04}$ \\ \cline{2-5}
        ~ & $0.4$ & $2.75\pm 0.96$ & $\mathbf{0.23\pm 0.02}$ & $0.72\pm 0.03$ \\ \hline
    \end{tabular}

\end{table}

\begin{table}[htb]
    \caption{Experimental results on the genomics dataset applying oversampling (OV) with regularization coefficient $C=10^{-4}$ and C-crit $= 0.3$. OV stands for the number of additional negative bags in oversampling.}
    \label{tab:oversamp}
    \centering
    \begin{tabular}{c|c||c|c|c}
        
        \hline
         Method & OV & CL & HL$^{\downarrow}$ & AP$^{\uparrow}$ \\ \hline
        \multirow{3}*{MIMLSVM} & $0$ & / & $0.26\pm 0.03$ & $0.69\pm 0.06$ \\ \cline{2-5}
        ~ & $100$ & / & $0.27\pm 0.04$ & $0.69\pm 0.57$ \\ \cline{2-5}
        ~ & $200$ & / & $0.30\pm 0.04$ & $0.71\pm 0.04$ \\ \hline
        \multirow{3}*{Ours} & $0$ & $5\pm 4.58$ & $0.22\pm 0.03$ & $0.67\pm 0.04$ \\ \cline{2-5}
        ~ & $100$ & $10.6\pm 10.14$ & $0.22\pm 0.01$ & $0.72\pm 0.07$ \\ \cline{2-5}
        ~ & $200$ & $5.20\pm 4.60$ & $\mathbf{0.20\pm 0.03}$ & $\mathbf{0.74\pm 0.09}$ \\ \hline
    \end{tabular}

\end{table}

On the genomics dataset, we compared the performance of two major criteria, namely Hamming Loss (HL) and Average Precision (AL) between our method and naive single-instance single-label learning(SISL) method, as well as MIMLSVM. As demonstrated in Table~\ref{fix_c}, given the dataset with potential noises in biomarker identification, conventional methods are likely to have high bias. SISL is likely of this case, where instance-level SVMs are learned given the limited instance-level examples, then the bag-level relevance is decided when at least one instance in the bag is predicted to be positive, which is likely to overfit the instance-level data. Although MIMLSVM performs inference at bag level, it still cannot capture label correlations in the multi-label data. On the contrary, our method benefits from constructive clustering at bag level as well as the chained modelling of the label correlations, observing that trained model with a longer chain (CL) has better performance on HL and AP.

Our method was further elevated when applying oversampling strategy as stated in Section~\ref{subsec:oversample}. As shown in Table~\ref{tab:oversamp}, comparing to the situation without oversampling (OV$=0$), oversampling further improves the performance of our method, especially in AP, where the score increased by around 7\%. The testing data doesn't include any manually inserted negative instances or bags, proving that oversampling indeed helps generate more reliable decision boundaries.

\section{CONCLUSIONS}

In this paper, we present a multi-instance multi-label framework for gene mutation detection on Hepatocellular carcinoma, which not only mitigates the influence of irrelevant instances, but also leverages the relationships among gene mutations. Besides, an oversampling strategy is applied to assist our MIML framework to overcome the imbalance issue of the dataset. Extensive experimental results demonstrate the effectiveness of the proposed MIML framework. In our future work, the correlation between image appearance and gene mutations will be explored by using MIML.


\bibliographystyle{IEEEbib}
\bibliography{refs.bib}
\end{document}